\documentclass[10pt,twocolumn,letterpaper]{article}
\usepackage{cvpr}

\usepackage{graphicx}
\usepackage{amsmath}
\usepackage{amssymb}
\usepackage{booktabs}
\usepackage{xcolor, color, colortbl}
\usepackage{multirow}
\usepackage{pifont}
\usepackage{array}
\usepackage[accsupp]{axessibility}

\usepackage[breaklinks,colorlinks]{hyperref}

\usepackage[capitalize]{cleveref}
\crefname{section}{Sec.}{Secs.}
\Crefname{section}{Section}{Sections}
\Crefname{table}{Table}{Tables}
\crefname{table}{Tab.}{Tabs.}

\newcommand{\net}{CompletionFormer}
\newcommand{\cmark}{\ding{51}}%
\newcommand{\xmark}{\ding{55}}%

\definecolor{red}{rgb}{0.95,0.4,0.4}
\definecolor{blue}{rgb}{0.4,0.4,0.95}
\definecolor{darkblue}{rgb}{0,0,0.8}
\definecolor{darkred}{rgb}{0.8,0,0}
\definecolor{darkgreen}{rgb}{0,0.5,0}
\definecolor{grey}{rgb}{0.6,0.6,0.6}
\definecolor{higher}{RGB}{232, 161, 148}
\definecolor{lower}{RGB}{148, 187, 232}

\definecolor{depthblue}{rgb}{0.0, 0.0, 1.0}
\definecolor{depthgreen}{rgb}{0.0, 1.0, 0.0}
\definecolor{crossyellow}{rgb}{1.0, 1.0, 0.0}

\begin{document}

\title{CompletionFormer: Depth Completion with Convolutions and \\ Vision Transformers}

\author{Youmin Zhang$^{1}$ \hspace{1cm} Xianda Guo$^{2}$ \hspace{1cm} Matteo Poggi$^{1}$ \\
Zheng Zhu$^{2}$ \hspace{1cm} Guan Huang$^{2}$ \hspace{1cm} Stefano Mattoccia$^{1}$ \\ \vspace{-0.3cm} \\
$^{1}$ University of Bologna \hspace{0.3cm} $^{2}$ PhiGent Robotics \\
$^{1}${\tt\small \{youmin.zhang2, m.poggi, stefano.mattoccia\}@unibo.it} \\ 
}

\maketitle

\begin{abstract}

Given sparse depths and the corresponding RGB images, depth completion aims at spatially propagating the sparse measurements throughout the whole image to get a dense depth prediction. Despite the tremendous progress of deep-learning-based depth completion methods, the locality of the convolutional layer or graph model makes it hard for the network to model the long-range relationship between pixels. While recent fully Transformer-based architecture has reported encouraging results with the global receptive field, the performance and efficiency gaps to the well-developed CNN models still exist because of its deteriorative local feature details.  
This paper proposes a Joint Convolutional Attention and Transformer block (JCAT), which deeply couples the convolutional attention layer and Vision Transformer into one block, as the basic unit to construct our depth completion model in a pyramidal structure. This hybrid architecture naturally benefits both the local connectivity of convolutions and the global context of the Transformer in one single model. As a result, our CompletionFormer outperforms state-of-the-art CNNs-based methods on the outdoor KITTI Depth Completion benchmark and indoor NYUv2 dataset, achieving significantly higher efficiency (nearly 1/3 FLOPs) compared to pure Transformer-based methods. Code is available at \url{https://github.com/youmi-zym/CompletionFormer}.

\end{abstract}

\section{Introduction}

\begin{figure*}[ht]
    \centering
    
    \includegraphics[width=2.0\columnwidth]{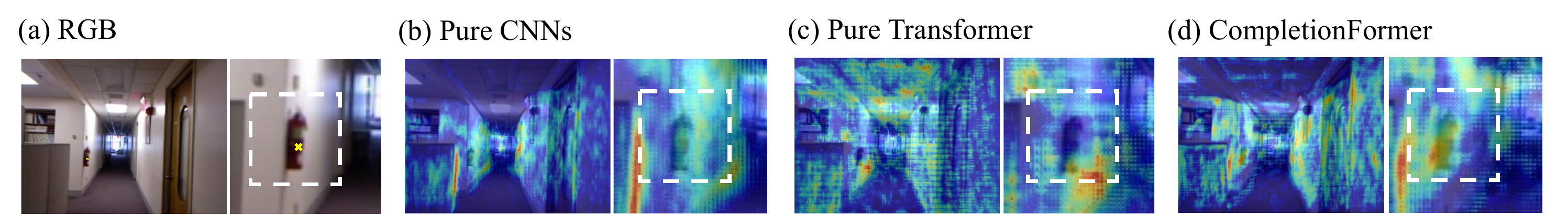}
    \vspace{-0.2cm}
    \caption{\textbf{Comparison of attention maps of pure CNNs, Vision Transformer, and the proposed \net{} with joint CNNs and Transformer structure.} The pixel highlighted with a yellow cross in RGB image (a) is the one we want to observe how the network predicts it. Pure CNNs architecture (b) activates discriminative local regions (\ie{}, the region on the fire extinguisher), whereas pure Transformer based models (c) activate globally yet fail on local details. In contrast, our full \net{} (d) can retain both the local details and global context.
    }
    
    \label{fig:attention}
\end{figure*}

Active depth sensing has achieved significant gains in performance and demonstrated its utility in numerous applications, such as autonomous driving and augmented reality. Although depth maps captured by existing commercial depth sensors (\eg, Microsoft Kinect~\cite{kinect}, Intel RealSense~\cite{realsense}) or depths points within the same scanning line of LiDAR sensors are dense, the distance between valid/correct depth points could still be far owing to the sensor noise, challenging conditions such as transparent, shining, and dark surfaces, or the limited number of scanning lines of LiDAR sensors. To address these issues, depth completion~\cite{CSPN++,DySPN,NLSPN,GuideFormer}, which targets at completing and reconstructing the whole depth map from sparse depth measurements and a corresponding RGB image (\ie, RGBD), has gained much attention in the latest years.

For depth completion, one key point is to get the depth affinity among neighboring pixels so that reliable depth labels can be propagated to the surroundings~\cite{CSPN++,CSPN,PENet,DySPN,NLSPN}. Based on the fact that the given sparse depth could be highly sparse due to noise or even no measurement being returned from the depth sensor, it requires depth completion methods to be capable of 1) detecting depth outliers by measuring the spatial relationship between pixels in both local and global perspectives; 2) fusing valid depth values from close or even extremely far distance points. All these properties ask the network for the potential to capture both local and global correlations between pixels. Current depth completion networks collect context information with the widely used convolution neural networks (CNNs)~\cite{CSPN++,CSPN,PENet,FusionNet,DySPN,DeepLiDAR,NLSPN,RobustDC} or graph neural network~\cite{ACMNet,xiong2020sparse}. However, both the convolutional layer and graph models can only aggregate within a local region, \eg{} square kernel in $3\times 3$ for convolution and kNN-based neighborhood for graph models~\cite{ACMNet,xiong2020sparse}, making it still tough to model global long-range relationship, in particular within the shallowest layers of the architecture. Recently, GuideFormer~\cite{GuideFormer} resorts fully Transformer-based architecture to enable global reasoning. Unfortunately, since Vision Transformers project image patches into vectors through a single step, this causes the loss of local details, resulting in ignoring local feature details in dense prediction tasks~\cite{ConFormer,coat}. For depth completion, the limitations affecting pure CNNs or Transformer based networks also manifest, as shown in \cref{fig:attention}. Despite \textit{any} distance the reliable depth points could be distributed at, exploring an elegant integration of these two distinct paradigms, \ie{} CNNs and Transformer, has not been studied for depth completion yet.

\begin{figure*}[t]
	\centering
		\includegraphics[width=1.6\columnwidth]{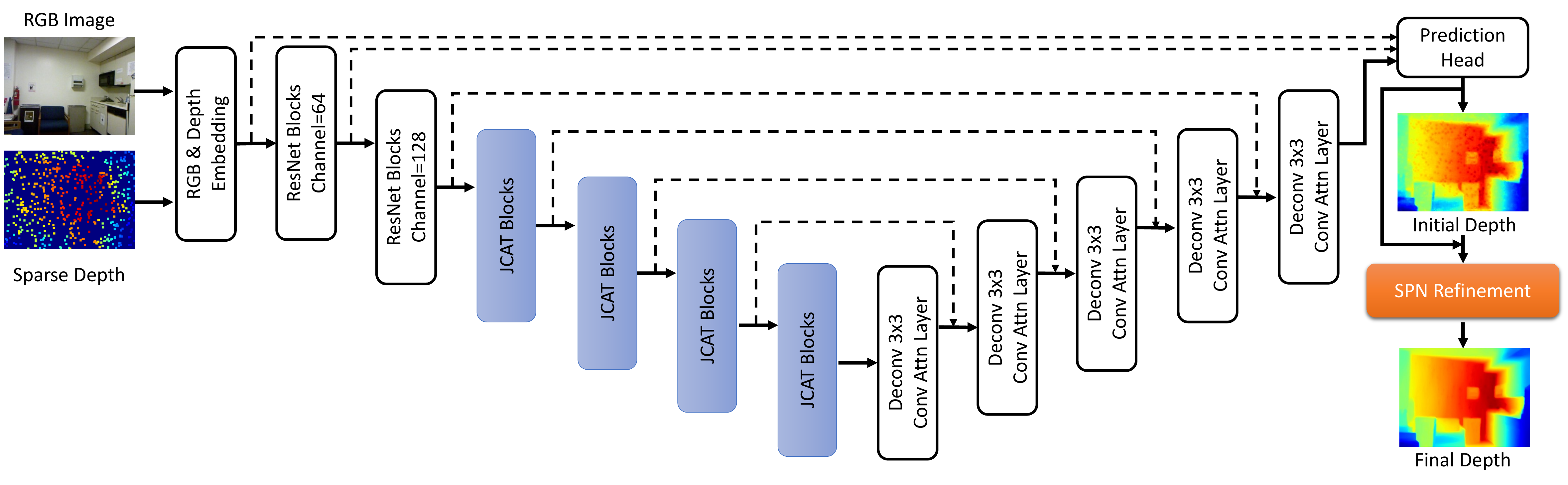}
		\caption{\textbf{\net{} Architecture.} Given the sparse depth and corresponding RGB image, a U-Net backbone strengthened with JCAT block is used to perform the depth and image information interaction at multiple scales. Features from different stages are fused at full resolution and fed for initial prediction. Finally, a spatial propagation network (SPN) is exploited for final refinement.
		} 
	\label{fig:framework}
\end{figure*}

In this work, we propose \net{}, a pyramidal architecture coupling CNN-based local features with Transformer-based global representations for enhanced depth completion. Generally, there are two gaps we are facing: 1) the content gap between RGB and depth input; 2) the semantic gap between convolution and Transformer. As for the multimodal input, we propose embedding the RGB and depth information at the early network stage. Thus our \net{} can be implemented in an efficient single-branch architecture as shown in \cref{fig:framework} and multimodal information can be aggregated throughout the whole network. Considering the integration of convolution and Transformer, previous work has explored from several different perspectives~\cite{mpvit,coat,ConFormer,CMT,integrationvit} on image classification and object detection. Although state-of-the-art performance has been achieved on those tasks, high computation cost~\cite{mpvit} or inferior performance~\cite{mpvit,CMT} are observed when these networks are directly adapted to depth completion task. To promise the combination of self-attention and convolution still being efficient, and also effective,  we embrace convolutional attention and Transformer into one block and use it as the basic unit to construct our network in a multi-scale style. Specifically, the Transformer layer is inspired by Pyramid Vision Transformer~\cite{pvt}, which adopts spatial-reduction attention to make the Transformer layer much more lightweight. As for the convolution-related part, the common option is to use plain convolutions such as Inverted Residual Block~\cite{mobilenetv2}. However, the huge semantic gap between convolution and the Transformer and the lost local details by Transformer require the convolutional layers to increase its own capacity to compensate for it. Following this rationale, we further introduce spatial and channel attention~\cite{cbam} to enhance convolutions. As a result, without any extra module to bridge the content and semantic gaps~\cite{ConFormer,mpvit,GuideFormer}, every convolution and Transformer layer in the proposed block can access the local and global features. Hence, information exchange and fusion happen effectively at every block of our network.

To summarize, our main contributions are as follows:
\begin{itemize}
    \item We propose integrating Vision Transformer with convolutional attention layers into one block for depth completion, enabling the network to possess both local and global receptive fields for multi-modal information interaction and fusion. In particular, spatial and channel attention are introduced to increase the capacity of convolutional layers.
    \item Taking the proposed \underline{J}oint \underline{C}onvolutional \underline{A}ttention and \underline{T}ransformer (JCAT) block as the basic unit, we introduce a single-branch network structure, \ie{} \net{}. This elegant design leads to a comparable computation cost to current CNN-based methods while presenting significantly higher efficiency when compared with pure Transformer based methods.
    \item Our \net{} yields substantial improvements to depth completion compared to state-of-the-art methods, especially when the provided depth is \textit{very} sparse, as often occurs in practical applications.
\end{itemize}

\section{Related Work}

\textbf{Depth Completion.} Scene depth completion has become a fundamental task in computer vision with the emergence of active depth sensors. Recently, following the advance of deep learning, fully-convolutional network has been the prototype architecture for current state-of-the-art on depth completion. Ma \emph{et al.}~\cite{Ma2017SparseToDense,ma2019self} utilize a ResNet~\cite{resnet} based encoder-decoder architecture, \ie{} U-Net, within either a supervised or self-supervised framework to predict the dense output. 
To preserve the accurate measurements in the given sparse depth and also perform refinement over the final depth map, CSPN~\cite{CSPN} appends a convolutional spatial propagation network (SPN~\cite{spn}) at the end of U-Net to refine its coarse prediction. Based on CSPN, learnable convolutional kernel sizes and a number of iterations are proposed to improve the efficiency~\cite{CSPN++}, and the performance could be further improved by using unfixed local neighbors~\cite{DSPN,NLSPN} and independent affinity matrix for each iteration~\cite{DySPN}. For all these SPN-based methods, while a larger context is observed within recurrent processing, the performance is limited by the capacity of the convolutional U-Net backbone. Accordingly, we strengthen the expressivity of the U-Net backbone with local and global coherent context information, proving effective in improving performance.

Rather than depending on a single branch, multi-branch networks~\cite{FusionNet,FCFRNet,PENet,GuideNet,DeepLiDAR,semattnet,singlergbd} are also adopted to perform multi-modal fusion. The common way to fuse the multi-modal information is simple concatenation or element-wise summation operation. More sophisticated strategies like image-guided spatially-variant convolution~\cite{GuideNet,RigNet}, channel-wise canonical correlation analysis~\cite{zhong2019deep},
neighbour attention mechanism~\cite{neighborattention} and attention-based graph propagation~\cite{ACMNet,xiong2020sparse} were also proposed to enhance local information interaction and fusion. Instead of pixel-wise operation or local fusion, recently, GuideFormer~\cite{GuideFormer} proposed a dual-branch fully Transformer-based network to embed the RGB and depth input separately, and an extra module is further designed to capture inter-modal dependencies. The independent design for each input source leads to huge computation costs (near 2T FLOPs with the $352\times 1216$ input). In contrast, our \net{} in one branch brings significant efficiency (559.5G FLOPs), and the included convolutional attention layer complements the disadvantage of a Transformer in local details. 

\textbf{Vision Transformer.}
Transformers~\cite{mpvit,swin} are first introduced in natural language processing~\cite{nlp}, then also showing great potential in the fields of image classification~\cite{vit}, object detection~\cite{swin,mpvit,coat} and semantic segmentation~\cite{segformer}. Tasks related to 3D vision have also benefited from the enriched modeling capability of Transformer, such as stereo matching~\cite{sttr,CREStereo}, supervised~\cite{dpt,depthformer} and unsupervised monocular depth estimation~\cite{monovit}, optical flow~\cite{craft,GMA} and also depth completion~\cite{GuideFormer}. Instead of relying on pure Vision Transformer~\cite{GuideFormer}, in this paper, we explore the combination of Transformer and convolution into one block for depth completion. Compared to the general backbone networks (\eg{} ResNet~\cite{resnet} with fully CNN-based design, Swin Transformer~\cite{swin} and PVT~\cite{pvt} based on pure Transformer, MPViT~\cite{mpvit} and CMT~\cite{CMT} using both the convolutions and Vision Transformer), our proposed joint convolutional attention and Transformer block achieves much higher efficiency and performance on public benchmarks~\cite{NYUV2,KITTI}

\begin{figure*}[t]
	\centering
		\includegraphics[width=1.8\columnwidth]{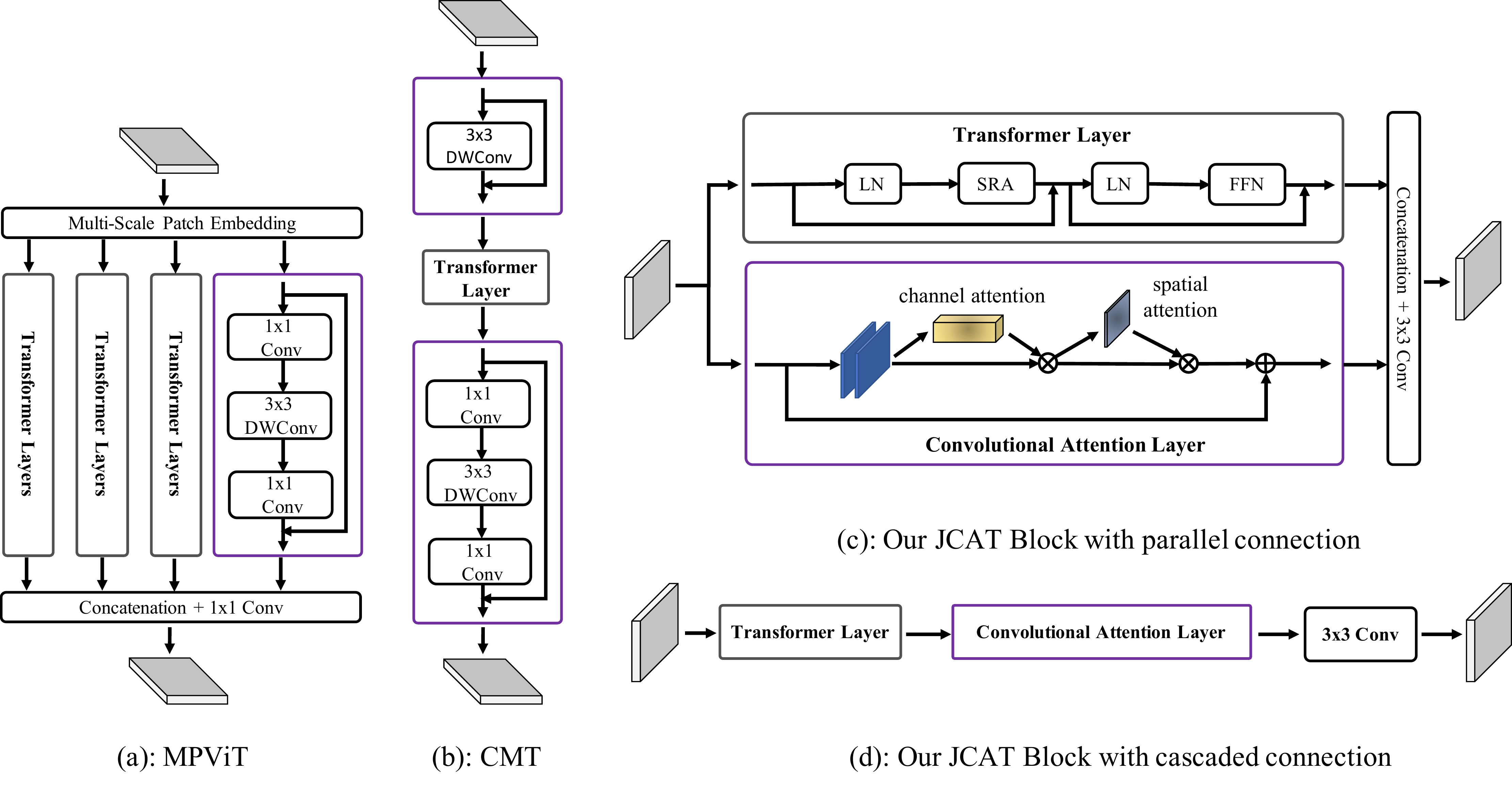}
		\caption{\textbf{Example of architecture with convolutions and Vision Transformer.} (a) Multi-Path Transformer Block of MPViT~\cite{mpvit}. (b) CMT Block of CMT-S~\cite{CMT}. (c) Our proposed JCAT block which contains two parallel streams, \ie{} convolutional attention and Transformer layer respectively. (d) The variant of our proposed block with cascaded connection.
		} 
	\label{fig:transformer_block}
\end{figure*}
\section{Method}

In practical applications, depth maps captured by sensors present various levels of sparsity and noise. Our goal is to introduce both the local features and global context information into the depth completion task so that it can gather reliable depth hints from any distance. The overall diagram of our \net{} is shown in ~\cref{fig:framework}. After obtaining depth and RGB image embedding, a backbone constructed by our JCAT block is used for feature extraction at multiple scales and the decoder provides full-resolution features for initial depth prediction. Finally, for the purpose of preserving accurate depth from the sparse input, we refine the initial estimation with a spatial propagation network.

\subsection{RGB and Depth Embedding}
\label{sec:embedding_module}
For depth completion, multimodal information fusion at an early stage has several advantages, 1) it makes the feature vector of each pixel possess both the RGB and depth information so that pixels with invalid depth still have a chance to be corrected by reliable depth measurements according to appearance similarity; 2) only one branch is required for the following network, which enables much efficient implementation. Therefore, we firstly use two separate convolutions to encode the input sparse depth map $\mathnormal{S}$ and RGB image $\mathnormal{I}$. The outputs are concatenated and further processed by another convolution layer to get the raw feature containing contents from both sources.

\subsection{Joint Convolutional Attention and Transformer Encoder}
It has been extensively studied how to build connections between pixels to implement depth propagation from reliable pixels while avoiding incorrect ones. Recently, convolution layer~\cite{CSPN++,CSPN,PENet,FusionNet,DySPN,DeepLiDAR,NLSPN,RobustDC} or attention based graph propagation~\cite{ACMNet,xiong2020sparse} has been the dominant operation for this purpose. Although a fully Transformer-based network~\cite{GuideFormer} has also been adopted for this purpose, it shows worse results and much higher computational cost compared to pure CNNs-based methods. Considering the complementary properties of these two-style operations, an elegant integration of these two paradigms is highly demanded for depth completion task. On the other hand, for classification and object detection tasks, MPViT~\cite{mpvit} and CMT~\cite{CMT} are two representative state-of-the-art networks on the combination of self-attention and convolution as shown in \cref{fig:transformer_block} (a) and (b) respectively. Generally, the integration can be implemented in a parallel or cascaded manner. Accordingly, inspired by their design, within \net{}, we propose a joint design as shown in~\cref{fig:transformer_block} (c) and (d). To decrease computation overhead but also get highly accurate depth completion result, our \net{} only contains single rather than multiple time-consuming Transformer-based paths as in MPViT~\cite{mpvit}. Furthermore, the representation power of convolution-based path is enhanced with spatial and channel attention.

\begin{table}[t]
    \centering
    \resizebox{0.9\columnwidth}{!}{
    \begin{tabular}{c|cccc}
    \toprule
    \net{} & \#Layers & Params (M) & FLOPs (G) \\
    \midrule
    Tiny & [2, 2, 2, 2] & 41.5 & 191.7 \\
    Small & [3, 3, 6, 3] & 78.3 & 231.8 \\
    Base & [3, 3, 18, 3] & 142.4 & 301.9 \\
    \bottomrule
    \end{tabular}
    }
    \caption{\textbf{\net{} Configurations.} \#Layers denotes the number of our JCAT blocks in each stage. For all model variants, the channels of 4 stages are 64, 128, 320, 512, respectively. FLOPs are measured using $480\times640$ input image.}
    \label{tab:configuration}
\end{table}

Specifically, our encoder has five stages, allowing features representation at different scales to communicate with each other effectively. In the first stage, to decrease the computation cost and memory overhead introduced by the Transformer layer, we use a series of BasicBlocks from ResNet34~\cite{resnet} to process and finally get downsampled feature map $F_{1}$ at half resolution. For the next four stages, we introduce our proposed JCAT block as the basic unit for framework design. 

Basically, for each stage $i \in \{2, 3, 4, 5\}$, it consists of a patch embedding module and $L_{i}$ repeated JCAT blocks. The patch embedding module firstly divides the feature map $F_{i-1}$ from previous stage $i-1$ into patches with size $2\times 2$. We implement it with a $3\times 3$ convolution layer and stride set to 2 as~\cite{pvt}, so it actually halves resolution for features $F_{i-1}$ and thus allows for obtaining a features pyramid $\{ F_{2}, F_{3}, F_{4}, F_{5} \}$, whose resolutions are $\{ 1/4, 1/8, 1/16, 1/32 \}$ with respect to the input image. Furthermore, position embedding is also included in the embedded patches and passed through the JCAT blocks.

\textbf{Joint Convolutional Attention and Transformer Block.} 
In overview, our JCAT block can be organized in a parallel or cascaded manner as shown in \cref{fig:transformer_block} (c) and (d) respectively. The Transformer layer is implemented in an efficient way as in Pyramid Vision Transformer~\cite{pvt}, which contains a spatial-reduction attention (SRA) layer with multi-head mechanism and a feed-forward layer (FNN). Given input features $F\in \mathbb{R}^{H_{i} \times W_{i} \times C}$ from the patch embedding module or last joint block (with $H_{i}$ and $W_{i}$ height and width of features at stage $i$, and $C$ number of channel), we firstly normalize it with layer normalization~\cite{layernorm} (LN) and then flatten it into vector tokens $X \in \mathbb{R}^{N\times C}$, where $N$ is the number of tokens and equals to $H_{i} \times W_{i}$, \ie{} the number of all pixels in the $F$. Using learned linear transformations $W^{Q}$, $W^{K}$, and $W^{V}\in\mathbb{R}^{C\times C}$, tokens $X$ are projected into corresponding query $Q$, key $K$, and value vectors $V\in \mathbb{R}^{N\times C}$. Here, the spatial scale of $K$ and $V$ is further reduced to decrease memory consumption, and then self-attention is performed as:

\begin{equation}
    \text{Attention}(Q, K, V) = \text{Softmax}(\frac{QK^{T}}{\sqrt{C_{head}}})V, 
    \label{eq: self_attention}
\end{equation}
with $C_{head}$ the channel dimension of each attention head in SRA. According to \cref{eq: self_attention}, each token in the entire input space $F$ is matched with any tokens, including itself. Our depth completion network benefits from the self-attention mechanism in two folds: 1) it extends the receptive field of our network to the full image in each Transformer layer; 2) as we have embedded each token with both depth and RGB image information, the self-attention mechanism explicitly compare the similarity of each pixel not only by appearance, but also by depth with dot-product operation. Thus, reliable depth information can be broadcasted to the whole image, enabling to correct erroneous pixels.

We boost the representation power of the convolutional path with channel and spatial attention~\cite{cbam}. On the one hand, it helps to model locally accurate attention and reduce noise. On the other hand, due to the semantic gap between convolution and Transformer, the increased modeling capacity by using the attention mechanism enables this path to focus on important features provided by the Transformer layer while suppressing the unnecessary ones. Finally, by concatenating the reshaped feature from the Transformer-based path, we fuse the two paths with a $3\times 3$ convolution and send it to the next block or stage.

Taking the proposed JCAT block as the basic unit, we build stages 2-5 with repeated configurations. As reported in \cref{tab:configuration}, we scale-up the 4 stages in \net{} from tiny, small to base scale. 
Our results demonstrate the superiority of JCAT design compared to recent Vision Transformers~\cite{mpvit,swin,pvt} in depth completion task.

\subsection{Decoder}
\label{sec:decoder}
In the decoder, outputs from each encoding layer are concatenated and further processed by the corresponding decoding layers via skip connections. To accommodate diverse scale features better, the features from previous decoder layer is upsampled to current scale with a deconvolution layer and convolutional attention mechanism~\cite{cbam} is also exploited to strengthen feature fusion in channel and spatial dimensions. Finally, the fused result from the decoder is concatenated with features from stage one and fed to the first convolution layer of the prediction head. Its output is concatenated with the raw feature from RGB and depth embedding module (\cref{sec:embedding_module}) and sent to another convolution, which is in charge of initial depth prediction $\mathnormal{D}^{0}$.

\subsection{SPN Refinement and Loss Function}
Considering that the accurate depth values from the sparse input may not be well preserved after going though U-Net~\cite{CSPN,PENet}, spatial propagation network~\cite{spn} has been a standard operation for final refinement. Recent work~\cite{CSPN,CSPN++,PENet,NLSPN} mainly focuses on improving the spatial propagation network from fixed-local to nonlocal propagation. While in our experiments (\cref{tab: ablation_components}), we observe that, with our enhanced U-Net backbone, the network is able to provide good depth affinity and thus obtain almost the same accuracy with fixed-local~\cite{CSPN,CSPN++} or non-local~\cite{NLSPN} neighbours for spatial propagation. With regard to CSPN++~\cite{CSPN++} consumes more computation cost, we adopt the non-local spatial propagation network~\cite{NLSPN} (NLSPN) for further refinement. Specifically, let $\mathnormal{D}^{t}=(d^{t}_{u,v}) \in \mathbb{R}^{H\times W}$ denotes the 2D depth map updated by spatial propagation at step $t$, where $d^{t}_{u,v}$ denotes the depth value at pixel $(u,v)$, and $H, W$ denotes the height and width of the $\mathnormal{D}^{t}$ respectively. The propagation of $d^{t}_{u,v}$ at step $t$ with its non-local neighbors $\mathnormal{N}_{u,v}^{NL}$ is defined as follows:

\begin{equation}
    \begin{aligned}
    d_{u,v}^{t} = w_{u,v}(0,0) d_{u,v}^{t-1} + \sum_{(i,j) \in \mathnormal{N}_{u,v}^{NL}, i\neq 0, j\neq 0} w_{u,v}(i,j) d_{i,j}^{t-1},
    \end{aligned}
    \label{eq:propagation}
\end{equation}
where $w_{u,v}(i,j) \in (-1, 1)$ describes the affinity weight between the reference pixel at $(u,v)$ and its neighbor pixel at $(i,j)$ and $w_{u,v}(0,0) =1 - \sum_{(i,j) \in \mathnormal{N}_{u,v}, i\neq 0, j\neq 0} w_{u,v}(i,j)$ stands for how much the original depth $d_{u,v}^{t-1}$ will be preserved. Moreover, the affinity matrix $w$ is also outputted by the decoder and modulated by a predicted confidence map from decoder to prevent less confident pixels from propagating into neighbors regardless of how large the affinity is.
After $K$ steps spatial propagation, we get the final refined depth map $\mathnormal{D}^{K}$. 

Finally, following~\cite{NLSPN}, a combined $L_{1}$ and $L_{2}$ loss is employed to supervise the network training as follows:

\begin{equation}
    \mathnormal{L}(\hat{\mathnormal{D}}, \mathnormal{D}^{gt}) = \frac{1}{|\mathnormal{V}|} \sum_{v\in\mathnormal{V}} \left( | \hat{\mathnormal{D}}_{v} - \mathnormal{D}_{v}^{gt} | + | \hat{\mathnormal{D}}_{v} - \mathnormal{D}_{v}^{gt} |^{2} \right),
\end{equation}
where $\hat{\mathnormal{D}} = \mathnormal{D}^{K}$, and $\mathnormal{V}$ is the set of pixels with valid depth in ground truth $\mathnormal{D}^{gt}$, and $|\mathnormal{V}|$ denotes the size of set $\mathnormal{V}$.

\section{Experiments}

\subsection{Datasets}

\textbf{NYUv2 Dataset~\cite{NYUV2}:} it consists of RGB and depth images captured by Microsoft Kinect~\cite{kinect} in 464 indoor scenes. Following the similar setting of previous depth completion methods~\cite{NLSPN,RobustDC}, our method is trained on 50,000 images uniformly sampled from the training set and tested on the 654 images from the official labeled test set for evaluation. For both training and test sets, the original frames of size $640\times 480$ are half down-sampled with bilinear interpolation and then center-cropped to $304\times228$. The sparse input depth is generated by random sampling from the dense ground truth. 

\textbf{KITTI Depth Completion (DC) Dataset~\cite{KITTI}:} it contains 86\,898 training data, 1\,000 selected for validation, and 1\,000 for testing without ground truth. The original depth map obtained by the Velodyne HDL-64e is sparse, covering about 5.9\% pixels. The dense ground truth is generated by collecting LiDAR scans from 11 consecutive temporal frames into one, producing near 30\% annotated pixels. These registered points are verified with the stereo image pairs to eliminate noisy values. Since there is no LiDAR return at the top of the image, following~\cite{NLSPN}, input images are bottom center-cropped to $240\times1216$ for training, validation and testing phases.

\subsection{Implementation Details}
We implement our model in PyTorch~\cite{pytorch} on 4 NVIDIA 3\,090 GPUs, using AdamW~\cite{adamw} as optimizer with an initial learning rate of 0.001, $\beta_{1}=0.9,\; \beta_{2}=0.999$, weight decay of 0.01. The batch size per GPU is set to 3 and 12 on KITTI DC and NYUv2 datasets, respectively. On the NYUv2 dataset, we train the model for 72 epochs and decay the learning rate by a factor of 0.5 at epochs 36, 48, 60, 72. For the KITTI DC dataset, the model is trained with 100 epochs, and we reduce the learning rate by half at epochs 50, 60, 70, 80, 90. The {supplementary material} outlines more details about network parameters.

\begin{table}[t]
    \centering
    \begin{tabular}{c}
    \resizebox{0.96\columnwidth}{!}{
    \begin{tabular}{ll|cc|cc}
    \toprule
    & \multirow{2}{*}{\net{}} & \cellcolor{lower} RMSE$\downarrow$ & \cellcolor{lower} MAE$\downarrow$ & Params.$\downarrow$ & FLOPs $\downarrow$ \\
    & & \cellcolor{lower}(mm) & \cellcolor{lower}(mm) & (M)  & (G) \\
    \midrule
    (A) & w/ cascaded connection & 91.5 & 35.7 & 82.6 & 429.6 \\
    (B) & w/ parallel connection & \textbf{90.0} & \textbf{35.0} & 82.6 & 429.6  \\
    \midrule
    (C) & w/o Spatial and Channel Attention & 91.1 & 35.5 & \textbf{82.5} & \textbf{429.4} \\
    \midrule
    (D) & w/ dual-branch encoders & 94.0 & 36.4 & 161.0 & 661.4  \\
    
    \bottomrule
    \end{tabular}
    }
    \\
    \resizebox{0.96\columnwidth}{!}{
    \begin{tabular}{llcc|cc|cc}
    \toprule
    & Backbone & Attention & Refinement & \cellcolor{lower} RMSE$\downarrow$ & \cellcolor{lower} MAE$\downarrow$ & Params.$\downarrow$ & FLOPs $\downarrow$  \\
    & Type & Decoder & Iterations & \cellcolor{lower}(mm) & \cellcolor{lower}(mm) & (M)  & (G) \\
    \midrule
    (E) & ResNet34~\cite{resnet} & \xmark & 18 & 92.3 & 36.1 & \textbf{26.4} & 542.2 \\
    (F) & ResNet34~\cite{resnet} & \cmark & 18 & 91.4 & 35.5 & 28.1 & 582.1  \\
    (G) & Swin-Tiny~\cite{swin} & \cmark & 18 & 92.6 & 36.4 & 38.1 & 634.8  \\
    (H) & PVT-Large~\cite{pvt} & \cmark & 18 & 91.4 & 35.6 & 68.3 & 419.8 \\
    (I) & MPViT-Base~\cite{mpvit} & \cmark & 18 & 91.0 & 35.5 & 83.1 & 1259.3 \\
    (J) & CMT-Base~\cite{CMT} & \cmark & 18 & 92.0 & 35.9 & 47.6 & \textbf{358.7}  \\
    (K) & Ours-Small & \cmark & 18 & 90.1 & 35.2 & 82.6 & 439.1 \\
    \midrule
    (L) & Ours-Small & \cmark & 6 & \textbf{90.0} & 35.0 & 82.6 & 429.6 \\ 
    (M) & Ours-Small & \cmark & CSPN++ & 90.3 & \textbf{34.9} & 82.7 & 446.4 \\
    \midrule
    (N) & Ours-Tiny & \cmark & 6 & 90.9 & 35.3 & 45.8 & 389.4 \\
    (O) & Ours-Base & \cmark & 6 & 90.1 & 35.1 & 146.7 & 499.6 \\

    \bottomrule
    \end{tabular}
    }
    \end{tabular}
    \vspace{-0.3cm}
    \caption{\textbf{Ablation study on NYU Depth v2~\cite{NYUV2}}.
    We ablate the settings of our network in the following aspects: the backbone type, the convolutional attention mechanism in decoder and the iterations of NLSPN Refinement module. FLOPs are measured with input resolution $480\times 640$.}\vspace{-0.3cm}
    \label{tab: ablation_components}
\end{table}

\begin{table}[t]
    \centering
    \resizebox{0.99\columnwidth}{!}{
		\begin{tabular}{l|c|c|c|c|c|c|c}
		\toprule
        & SPN & ResNet34 & Swin-Tiny & PVT-Large & MPViT-Base & CMT-Base & Ours-Small \\
        \hline
        \multirow{2}{*}{RMSE(mm)$\downarrow$} & \xmark & 106.5 & 106.5 & 106.7 & 100.2 & 108.4 & \textbf{99.2} \\
        & \cmark & 91.4 & 92.6 & 91.4 & 91.0 & 92.0 & \textbf{90.0} \\
		\bottomrule	
		\end{tabular}
	}
	\vspace{-0.3cm}
	\caption{\textbf{Ablation study without SPN module.} We report the accuracy of different backbone with/without SPN on NYUv2.}
	\vspace{-0.3cm}
	\label{table:no_spn}

\end{table}

\subsection{Evaluation Metrics}
Following the KITTI benchmark and existing depth completion methods~\cite{NLSPN,RobustDC}, given the prediction $\hat{\mathnormal{D}}$ and ground truth $\mathnormal{D}^{gt}$, we use the standard metrics for evaluation: (1) root mean square error (RMSE); (2) mean absolute error (MAE); (3) root mean squared error of the inverse depth (iRMSE); (4) mean absolute error of the inverse depth (iMAE); (5) mean absolute relative error (REL).

\subsection{Ablation Studies and Analysis}

We assess the impact of the main components of our \net{} on NYUv2 dataset~\cite{NYUV2}. Following previous methods~\cite{DySPN,NLSPN}, we randomly sample 500 depth pixels from the ground truth depth map and input them along with the corresponding RGB image for network training. Results are reported in  \cref{tab: ablation_components}.

\textbf{Cascaded vs Parallel Connection.}
We can notice that the cascaded design (A) shows inferior performance compared to parallel style (B). This conclusion is also confirmed in (I) and (J), as CMT-Base gets even worse RMSE (92.0) than MPViT-Base~\cite{mpvit} (91.0). It indicates the parallel connection is more suitable for information interaction between streams with different contents and semantics on depth completion task. Thus, we adopt parallel connection as our final scheme. 

\textbf{Spatial and Channel Attention in JCAT block.} Previous methods combine the Transformer with plain convolutions~\cite{mpvit,CMT,ConFormer,integrationvit}. Here, we also ablate the case when we disable the spatial and channel attention at the convolutional path of our proposed JCAT block (C). The drop in accuracy (RMSE increases from 90.0 to 91.1) confirms that increasing the capacity of convolutions is vital when combining convolution with Vision Transformer, and it only leads to negligible FLOPs increase (0.2G FLOPs).

\textbf{Single- or Dual-branch Encoder.} Similar to previous methods~\cite{GuideFormer,GuideNet}, we test the dual-branch architecture, which encodes the RGB and depth information separately (D). For feature communication between two branches, we include the spatial and channel attention mechanism~\cite{cbam} at the end of each stage in the encoder. However, the worse results compared to our single-branch design (B) demonstrates that embedding the multimodal information at the early stage is much more effective and efficient.

\textbf{Comparisons with General Feature Backbones.} In RMSE, our novel \net{} in small scale (K) outperforms pure CNNs based method (F) and those pure Transformer-based variants (G, H) counting comparable FLOPs with respect to our model. In particular, compared to the recent MPViT-Base~\cite{mpvit} (I) and CMT-Base~\cite{CMT} (J) which also integrate CNNs and Transformer for feature extraction, our network achieves higher accuracy and much lower computational overhead (429.6G FLOPs) than MPViT-Base (1259.3G FLOPS). 

\textbf{Decoder.} Compared to our baseline (E), \ie, NLSPN~\cite{NLSPN}, introducing spatial and channel attention for multiscale feature fusion in the decoder (F) further improves the results, RMSE drops from 92.3 to 91.4. 

\textbf{SPN Refinement Iterations.} Our network (L) requires as few as 6 iterations for SPN refinement to converge to the best result. Compared to baseline (E) which requires 18 iterations, our enhanced U-Net has collected information from the whole image and thus doesn't need large iterations to propagate to long distance. Even when the non-local refinement in NLSPN replaced with fixed-local neighbors in CSPN++~\cite{CSPN++} (M), the accuracy remains almost the same. It indicates that our \net{} can learn good affinity locally and globally, thus helping to soften the problem raised by the limited and fixed range of aggregation in CSPN++~\cite{CSPN++}.

\textbf{Model Scales.} Our models in various scales (L, N, O), benefiting from local and global cues, achieve significant improvement compared to the pure CNN-based baseline (E), while counting fewer FLOPs. To trade-off between accuracy and efficiency, we select our model in small scale (L) as our final architecture for the remaining experiments.

\textbf{With/without SPN refinement.} To conclude, in Tab. \ref{table:no_spn} we show the results achieved by different backbones with and without SPN refinement. Ours yields the most accurate results in both cases.

\begin{table}[t]
\centering
\resizebox{0.8\columnwidth}{!}{
\begin{tabular}{c|c|cccc}
    \toprule
    Scanning & \multirow{2}{*}{Method}
    & \cellcolor{lower}RMSE$\downarrow$ & \cellcolor{lower}MAE$\downarrow$ & \cellcolor{lower}iRMSE$\downarrow$ & \cellcolor{lower}iMAE$\downarrow$ \\
    Lines & & \cellcolor{lower}(mm) & \cellcolor{lower}(mm) & \cellcolor{lower}(1/km) & \cellcolor{lower}(1/km) \\
    \midrule
    \multirow{4}{*}{1} & 
    NLSPN & 3507.7 & 1849.1 & 13.8 & 8.9 \\
    & DySPN & 3625.5 & 1924.7 & 13.8 & 8.9 \\
    & Ours-ViT & 3507.2 & 1807.7 & 12.1 & 7.8 \\
    & Ours & \textbf{3250.2} & \textbf{1582.6} & \textbf{10.4} & \textbf{6.6} \\
    
    \hline
    
    \multirow{4}{*}{4} & 
    NLSPN & 2293.1 & 831.3 & 7.0 & 3.4 \\
    & DySPN & 2285.8 & 834.3 & 6.3 & 3.2 \\
    & Ours-ViT & 2241.2 & 795.9 & 5.8 & 2.9  \\
    & Ours & \textbf{2150.0} & \textbf{740.1} & \textbf{5.4} & \textbf{2.6} \\
    \hline
    
    \multirow{4}{*}{16} & 
    NLSPN & 1288.9 & 377.2 & 3.4 & 1.4 \\
    & DySPN & 1274.8 & 366.4 & 3.2 & 1.3 \\
    & Ours-ViT & 1268.9 & 360.7 & 3.3 & 1.3 \\
    & Ours & \textbf{1218.6} & \textbf{337.4} & \textbf{3.0} & \textbf{1.2} \\
    \hline
    
    \multirow{4}{*}{64} & 
    NLSPN & 889.4 & 238.8 & 2.6 & 1.0 \\
    & DySPN & 878.5 & 228.6 & \textbf{2.5} & 1.0 \\
    & Ours-ViT & 872.0 & 226.2 & \textbf{2.5} & 1.0 \\
    & Ours & \textbf{848.7} & \textbf{215.9} & \textbf{2.5} & \textbf{0.9}  \\
    
    \bottomrule
    \end{tabular}}
    \vspace{-0.3cm}
\caption{\textbf{Ablation study on scanning lines of LiDAR sensor on KITTI DC~\cite{KITTI} datasets.} Ours-ViT denotes that only the Transformer layer is enabled in our proposed block.}\vspace{-0.5cm}
\label{table:lidar_lines}
\end{table}

\begin{table}[t]
    \centering
    \resizebox{1.0\columnwidth}{!}{
    \begin{tabular}{lc|ccccc}
    \toprule
     & & \multicolumn{5}{c}{\cellcolor{lower}RMSE$\downarrow$ (m)} \\
    \multicolumn{2}{c|}{Method} & PackNet-SAN & GuideNet & NLSPN & Ours-ViT & Ours \\
    \midrule
    & 0 & - & - & 0.562 & 0.544 & \textbf{0.490}  \\
    \cline{2-7}
    Sample & 50 & - & - & 0.223 & 0.218 & \textbf{0.208} \\
    \cline{2-7}
    Number & 200 & 0.155 & 0.142 & 0.129 & 0.130 & \textbf{0.127} \\
    \cline{2-7}
    & 500 & 0.120 & 0.101 & 0.092 & 0.091 & \textbf{0.090} \\
    
    \bottomrule
    \end{tabular}
    }
    \vspace{-0.3cm}
    \caption{\textbf{Sparsity Studies on the NYUv2 Dataset.} Evaluation with 0, 50, 200 and 500 samples.}\vspace{-0.5cm}
    \label{tab:ablation_sparsity}
\end{table}

\subsection{Sparsity Level Analysis}

Depth maps captured by sensors such as Microsoft Kinect~\cite{kinect} and Velodyne LiDAR sensor are unevenly distributed and often contain outliers. To compare the effectiveness of our model with current state-of-the-art methods~\cite{packnetsan,GuideNet,NLSPN,DySPN} when dealing with this challenging input depth, we manually generate sparse data at different settings for network training and testing. As for noise, since captured depth maps are often corrupted by sensor noise or displacement between RGB camera and depth sensor~\cite{DeepLiDAR}, we do not add extra noise to the input depth for experiments.

\begin{table}[t]
    \centering
    \resizebox{1.0\columnwidth}{!}{
    \begin{tabular}{l|cccc|cc}
    \toprule
    \multirow{3}{*}{Method} & \multicolumn{4}{c|}{KITTI DC} & \multicolumn{2}{c}{NYUv2} \\
    \cline{2-6}
    & \cellcolor{lower}MAE$\downarrow$ & \cellcolor{lower}iMAE$\downarrow$ & \cellcolor{lower}iRMSE$\downarrow$ & \cellcolor{lower}RMSE$\downarrow$ &  \cellcolor{lower}RMSE$\downarrow$ & \cellcolor{lower} \\ 
    & \cellcolor{lower}(mm) & \cellcolor{lower}(1/km)& \cellcolor{lower}(1/km) & \cellcolor{lower}(mm) & \cellcolor{lower} (m) & \multirow{-2}{*}{ \cellcolor{lower} REL$\downarrow$} \\
    \midrule
    CSPN~\cite{CSPN} & 279.46 & 1.15 & 2.93 & 1019.64 & 0.117 & 0.016 \\ 
    DeepLiDAR~\cite{DeepLiDAR} & 226.50 & 1.15 & 2.56 & 758.38 & 0.115 & 0.022 \\
    GuideNet~\cite{GuideNet} & 218.83 & 0.99 & 2.25 & 736.24 & 0.101 & 0.015 \\
    NLSPN~\cite{NLSPN} & 199.59 & 0.84 & 1.99 & 741.68 & 0.092 & \textbf{0.012}  \\
    PENet~\cite{PENet} & 210.55 & 0.94 & 2.17 & 730.08 & - & - \\ 
    ACMNet~\cite{ACMNet} & 206.09 & 0.90 & 2.08 & 744.91 & 0.105 & 0.015 \\ 
    TWISE~\cite{TWISE} & 195.58 & 0.82 & 2.08 & 840.20 & 0.097 & 0.013  \\
    RigNet~\cite{RigNet} & 203.25 & 0.90 & 2.08 & 712.66 & \textbf{0.090} & 0.013  \\
    GuideFormer~\cite{GuideFormer} & 207.76 & 0.97 & 2.14 & 721.48 & - & -  \\
    DySPN~\cite{DySPN} & 192.71 &
    0.82 & \textbf{1.88} & {709.12} & \textbf{0.090} & \textbf{0.012} \\
    \hline
    Ours ($L_1$) & \textbf{183.88} & \textbf{0.80} & 1.89 & 764.87 & - & - \\
    Ours ($L_1$+$L_2$) & 203.45 & 0.88 & 2.01 & \textbf{708.87} & \textbf{0.090} & \textbf{0.012} \\
    \bottomrule
    \end{tabular}
    }
    \vspace{-0.3cm}
    \caption{\textbf{Quantitative evaluation on KITTI DC and NYUv2.}}
    \vspace{-0.3cm}
    \label{tab:benchmark}
\end{table}

\begin{figure*}[t]
	\centering
		\includegraphics[width=1.8\columnwidth]{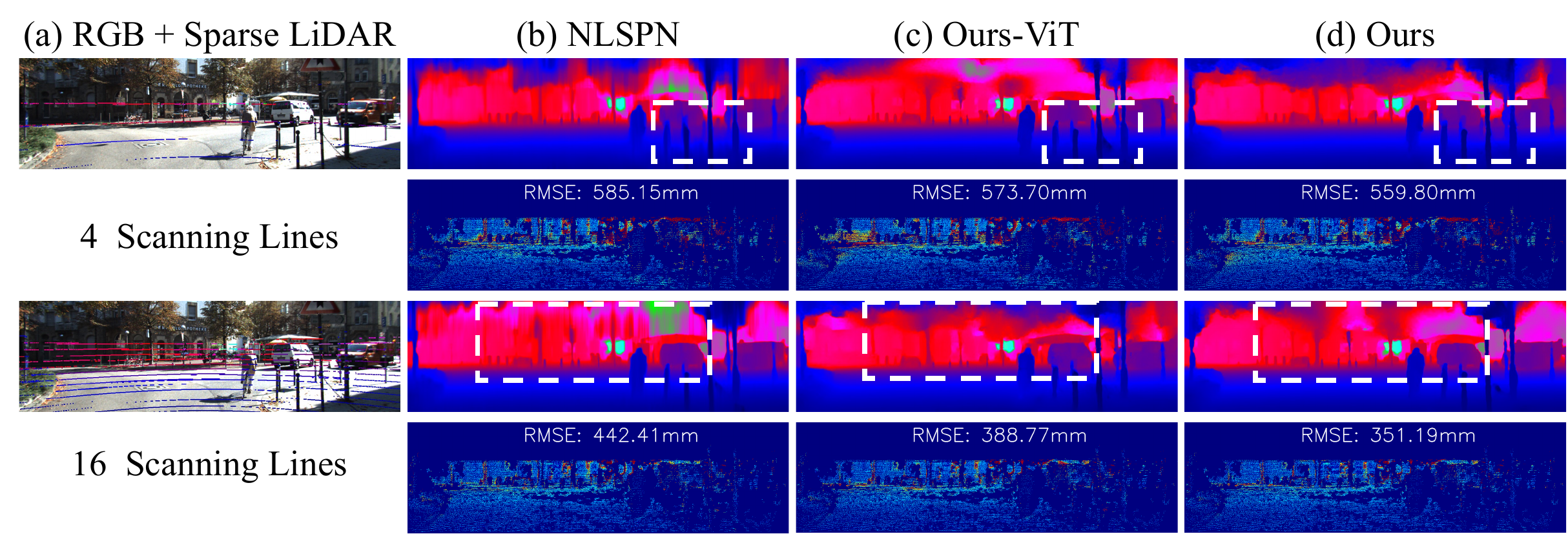}
		\vspace{-0.3cm}
		\caption{\textbf{Qualitative results on KITTI DC selected validation dataset with 4 and 16 LiDAR scanning lines.} We attach the subsampled LiDAR lines to the corresponding RGB image for better visualization. Ours-ViT denotes that only the Transformer layer is enabled in our proposed block. A colder color in depth and error maps denotes a lower value.
		} 
	\label{fig:kitti_sparse}
\end{figure*}

\begin{figure*}[!ht]
	\centering
		\includegraphics[width=1.8\columnwidth]{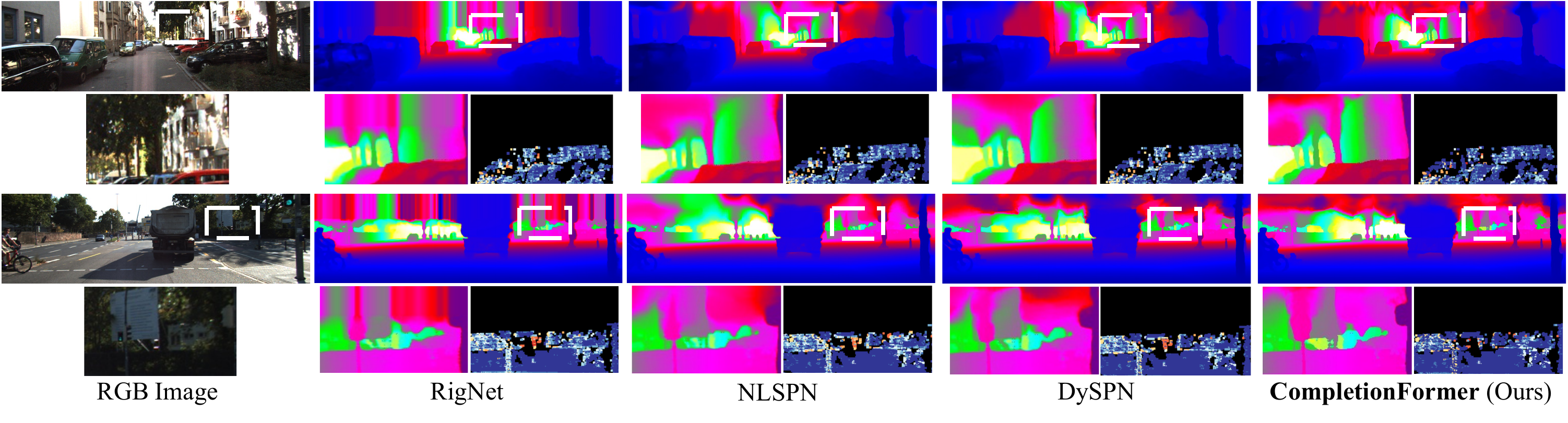}
		\vspace{-0.4cm}
		\caption{\textbf{Qualitative results on the KITTI depth completion test set.}  
		Comparisons of our method against state-of-the-art methods including RigNet~\cite{RigNet}, NLSPN~\cite{NLSPN}, DySPN~\cite{DySPN} are presented. 
		We provide RGB images, dense predictions, zoom-in views of challenging areas and corresponding error maps for better visualization. 
		} 
		\vspace{-0.3cm}

	\label{fig:kitti}
\end{figure*}

\textbf{Outdoor Scene.} We conduct exhaustive experiments on the KITTI DC dataset~\cite{KITTI} to illustrate the robustness of our CompletionFormer when input with different sparsity levels. For all experiments, we train on randomly selected 10\,000 RGB and LiDAR pairs from official training data (due to resource constraints) and test on the selected validation dataset provided by KITTI. Furthermore, following~\cite{TWISE}, we sub-sample the raw LiDAR points captured by Velodyne HDL-64e in azimuth-elevation space into 1, 4, 16 and 64 lines to simulate the LiDAR-like patterns. All methods have been fully \textbf{retrained} using the official code with variable lines of LiDAR inputs (For DySPN~\cite{DySPN}, as no public code available, the results have been provided to us by the author using the same selected training list). To get a throughout understanding of the local details encoded by convolution and global context gathered by Transformer, we present the results estimated by pure CNN-based methods (\ie{} NLSPN and DySPN), a fully Transformer-based variant of our network (Ours-ViT) and our \net{} complete architecture (Ours), which integrates both paradigms. As reported in \cref{table:lidar_lines}, thanks to the global correlations built by Transformers, our model relying on these blocks only (Ours-ViT) exhibits better results in all metrics as the LiDAR points get sparser. However, solely using Transformer layers makes it difficult to distinguish the objects from the background, as shown in \cref{fig:kitti_sparse}. By coupling the local features and global representations, our complete model (Ours) significantly decreases the errors in all metrics.

\textbf{Indoor Scene.} On the NYUv2 dataset~\cite{NYUV2}, we randomly sample 0, 50, 200 and 500 points from the ground truth depth map to mimic different depth sparsity levels while keeping the ground truth depth used for supervision unchanged. Both our model and NLSPN with publicly available code are {retrained} for a fair comparison, while the results of GuideNet~\cite{GuideNet} and PackNet-SAN~\cite{packnetsan} are taken from the original papers. In \cref{tab:ablation_sparsity}, \net{} with both CNNs and Transformers consistently outperforms all other methods in any cases. Qualitative results are provided in the {supplementary material}.

\subsection{Comparison with SOTA Methods}
This section comprehensively assesses the performance of state-of-the-art (SOTA) methods. On \textbf{indoor}, \ie{} NYUv2 dataset~\cite{NYUV2}, \net{} achieves the best results as reported in \cref{tab:benchmark}. When moving to \textbf{outdoor} dataset, MAE, iMAE, RMSE and iRMSE are adopted for benchmark on KITTI depth completion (DC) dataset~\cite{KITTI}. Empirically, our model trained with only $\mathnormal{L}_{1}$ loss achieves the best results on two among four metrics (MAE and iMAE). By jointly minimizing $L_1$ and $L_2$ losses, \net{} ranks first on RMSE metric among published methods. 
Qualitative results on the KITTI DC test dataset are provided in \cref{fig:kitti}. By integrating convolutions and Transformers, our model performs better near depth missing areas (\eg{} the zoom-in visualization on the second and fourth rows), textureless objects (\eg{} the cars on the first row) and small objects (\eg{} the pillars and tree stem far in the distance, on second and fourth rows).

\section{Conclusion and Limitations}

This paper proposed a single-branch depth completion network, \net{}, seamlessly integrating convolutional attention and Transformers into one block. Extensive ablation studies demonstrate the effectiveness and efficiency of our model in depth completion when the input is sparse. This novel design yields state-of-the-art results on indoor and outdoor datasets. Currently, \net{} runs at about 10 FPS: decreasing its runtime further to meet real-time requirements will be our future work.
\appendix

\section{Appendix}

\subsection{Qualitative Results on NYUv2 Dataset}
\begin{figure}[!ht]
	\centering
		\includegraphics[width=1.0\columnwidth]{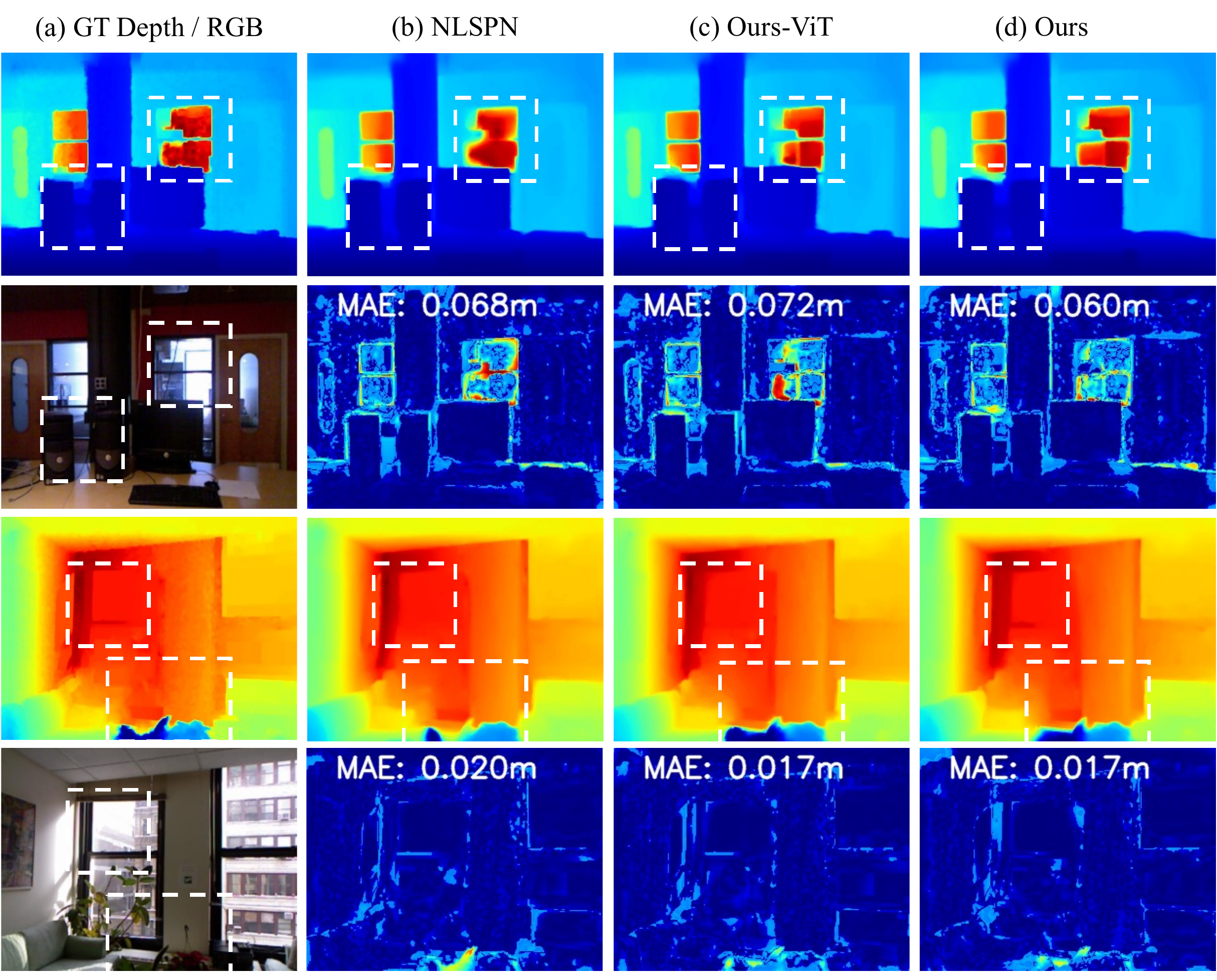}
		\caption{\textbf{Qualitative results on NYUv2 dataset.}  Comparisons of our method against state-of-the-art method, \ie{} NLSPN~\cite{NLSPN} are presented. We provide RGB images, and dense predictions. The colder the colors of the error map, the lower the errors. Ours-ViT denotes that only the Transformer layer is enabled in our proposed block.} 
	\label{fig:nyuv2}
\end{figure}

Qualitative results concerning the NYUv2 dataset~\cite{NYUV2} are provided in \cref{fig:nyuv2}. In both visualized cases, we can notice the improved results yielded by our \net{} compared to NLSPN~\cite{NLSPN}. Especially for the transparent regions near the windows in both cases, with local details of convolution and global cues of Transformer, our complete model (Ours) predicts clear object boundaries while NLSPN and Ours-ViT give blurry estimations.

\subsection{Model Architecture Details}
To better understand our architecture and to ease reproducibility, we present the network parameters of our \net{} in \cref{tab: network parameters}. 

\begin{table}[t]
	\begin{center}
        \resizebox{1.0\columnwidth}{!}{
			\begin{tabular}{c|c|c}
				\toprule
				 Name & Layer setting & Output dimension \\

				 \midrule
				 
				 \multicolumn{3}{c}{\textbf{RGB and Depth Embedding}} \\
				 
				 \midrule
				 
				 input &  & $ \begin{array}{c} \textrm{RGB   Image: }\;\; H \times W \times 3 \\ \textrm{Sparse Depth: } H \times W \times 1 \end{array} $ \\
				 \hline
				 conv\_separate & $ \begin{array}{c} \textrm{Conv} \; 3\times 3, 48 \;\; \textrm{ for RGB   Image} \\ \textrm{Conv}\; 3\times 3, 16 \textrm{ for Sparse Depth}  \end{array} $ & $ \begin{array}{c} \textrm{RGB   Feature: }\quad\quad\quad H \times W \times 48 \\ \textrm{Sparse Depth Feature: } H \times W \times 16 \end{array} $ \\
				 \hline
				 conv1 & $ \begin{array}{c} \textrm{concat [RGB} \textrm{,  Sparse Depth Feature]} \\ \textrm{Conv} 3\times 3, 64   \end{array} $ & $H \times W \times 64$ \\

				 \midrule
				 
				 \multicolumn{3}{c}{\textbf{Joint Convolutional Attention and Transformer Encoder}} \\
				 
				 \midrule
				 
				 conv2 & ResNet34~\cite{resnet} BasicBlock $\times 3$ & $H \times W \times 64$ \\
				 \hline
				 
				 conv3 & ResNet34~\cite{resnet} BasicBlock $\times 4$ & $ \frac{1}{2}H\times \frac{1}{2}W \times 128$ \\
				 \hline
				 
				 conv4 & Joint Convolutional Attention and Transformer Block $\times 3$ & $ \frac{1}{4}H\times \frac{1}{4}W \times 64$ \\ 
				 \hline
				 conv5 &Joint Convolutional Attention and Transformer Block $\times 3$  & $ \frac{1}{8}H\times \frac{1}{8}W \times 128$ \\ 
				 \hline
				 conv6 & Joint Convolutional Attention and Transformer Block $\times 6$  & $ \frac{1}{16}H\times \frac{1}{16}W \times 320$ \\ 
				 \hline
				 conv7 & Joint Convolutional Attention and Transformer Block $\times 3$  & $ \frac{1}{32}H\times \frac{1}{32}W \times 512$ \\ 
				 
				 \midrule
				 \multicolumn{3}{c}{\textbf{Decoder}} \\
				 
				 \midrule
				 
				 dec6 & $ \begin{array}{c} 
				 \textrm{ConvTranspose}\; 3\times 3, \; \textrm{stride}=2, \; 256 \\  \textrm{Convolutional Attention Layer} \\   
				 \end{array} $ & $ \frac{1}{16}H\times \frac{1}{16}W \times 256 $ \\
				 
				 \hline
				 
				 dec5 & $ \begin{array}{c} 
				 \textrm{concat [dec6, conv6]} \\
				 \textrm{ConvTranspose}\; 3\times 3, \; \textrm{stride}=2, \; 128 \\  \textrm{Convolutional Attention Layer} \\   
				 \end{array} $ & $ \frac{1}{8}H\times \frac{1}{8}W \times 128 $ \\
				 
				 \hline
				 
				 dec4 & $ \begin{array}{c} 
				 \textrm{concat [dec5, conv5]} \\
				 \textrm{ConvTranspose}\; 3\times 3, \; \textrm{stride}=2, \; 64 \\  \textrm{Convolutional Attention Layer} \\   
				 \end{array} $ & $ \frac{1}{4}H\times \frac{1}{4}W \times 64 $ \\
				 
				 \hline
				 
				 dec3 & $ \begin{array}{c} 
				 \textrm{concat [dec4, conv4]} \\
				 \textrm{ConvTranspose}\; 3\times 3, \; \textrm{stride}=2, \; 64 \\  \textrm{Convolutional Attention Layer} \\   
				 \end{array} $ & $ \frac{1}{2}H\times \frac{1}{2}W \times 64 $ \\
				 
				 \hline
				 
				 dec2 & $ \begin{array}{c} 
				 \textrm{concat [dec3, conv3]} \\
				 \textrm{ConvTranspose}\; 3\times 3, \; \textrm{stride}=2, \; 64 \\  \textrm{Convolutional Attention Layer} \\   
				 \end{array} $ & $ H\times W \times 64 $ \\
				 
				 \midrule
				 
				 \multicolumn{3}{c}{\textbf{Initial Depth, Confidence, Non-local Neighbors, Affinity Prediction Head}} \\
				 
				 \midrule
				 
				 dec1 & $ \begin{array}{c} \textrm{concat [dec2, conv2]} \\  \textrm{Conv}\; 3\times 3, 64   \end{array}$ & $H \times W \times 64$ \\ 
				 \hline
				 dec0 & $ \begin{array}{c} \textrm{concat [dec1, conv1]} \\  \textrm{Conv}\; 3\times 3, \eta   \end{array}$ & $H \times W \times \eta$ \\
				 
				 \midrule
				 
				 \multicolumn{3}{c}{\textbf{SPN Refinement}} \\
				 
				 \midrule
				 
				 refine & Spatial Propagation Network~\cite{NLSPN} with recurrent time $K=6$ & $H \times W \times 1$ \\
				 
				 \bottomrule	
			\end{tabular}
		}
	\end{center}
	\caption{Network parameters of \net{}. `concat' means performing concatenate operation at Channel dimension. For each prediction head, it takes almost the same design, and only the output channel $\eta$ is dependent on the output type, \eg{} $\eta=1$ for initial depth prediction.
	}
	\label{tab: network parameters}
\end{table}

\newpage

{\small
\bibliographystyle{ieee_fullname}
\bibliography{egbib}

\begin{thebibliography}{10}\itemsep=-1pt

\bibitem{layernorm}
Jimmy~Lei Ba, Jamie~Ryan Kiros, and Geoffrey~E Hinton.
\newblock Layer normalization.
\newblock {\em arXiv preprint arXiv:1607.06450}, 2016.

\bibitem{CSPN++}
Xinjing Cheng, Peng Wang, Chenye Guan, and Ruigang Yang.
\newblock Cspn++: Learning context and resource aware convolutional spatial
  propagation networks for depth completion.
\newblock In {\em AAAI}, volume~34, pages 10615--10622, 2020.

\bibitem{CSPN}
Xinjing Cheng, Peng Wang, and Ruigang Yang.
\newblock Depth estimation via affinity learned with convolutional spatial
  propagation network.
\newblock In {\em ECCV}, pages 103--119, 2018.

\bibitem{vit}
Alexey Dosovitskiy, Lucas Beyer, Alexander Kolesnikov, Dirk Weissenborn,
  Xiaohua Zhai, Thomas Unterthiner, Mostafa Dehghani, Matthias Minderer, Georg
  Heigold, Sylvain Gelly, et~al.
\newblock An image is worth 16x16 words: Transformers for image recognition at
  scale.
\newblock {\em ArXiv preprint arXiv:2010.11929}, 2020.

\bibitem{packnetsan}
Vitor Guizilini, Rares Ambrus, Wolfram Burgard, and Adrien Gaidon.
\newblock Sparse auxiliary networks for unified monocular depth prediction and
  completion.
\newblock In {\em CVPR}, pages 11078--11088, 2021.

\bibitem{CMT}
Jianyuan Guo, Kai Han, Han Wu, Yehui Tang, Xinghao Chen, Yunhe Wang, and Chang
  Xu.
\newblock Cmt: Convolutional neural networks meet vision transformers.
\newblock In {\em CVPR}, pages 12175--12185, June 2022.

\bibitem{resnet}
Kaiming He, Xiangyu Zhang, Shaoqing Ren, and Jian Sun.
\newblock Identity mappings in deep residual networks.
\newblock In {\em ECCV}, pages 630--645. Springer, 2016.

\bibitem{PENet}
Mu Hu, Shuling Wang, Bin Li, Shiyu Ning, Li Fan, and Xiaojin Gong.
\newblock Penet: Towards precise and efficient image guided depth completion.
\newblock In {\em ICRA}, pages 13656--13662. IEEE, 2021.

\bibitem{TWISE}
Saif Imran, Xiaoming Liu, and Daniel Morris.
\newblock Depth completion with twin surface extrapolation at occlusion
  boundaries.
\newblock In {\em CVPR}, pages 2583--2592, 2021.

\bibitem{GMA}
Shihao Jiang, Dylan Campbell, Yao Lu, Hongdong Li, and Richard Hartley.
\newblock Learning to estimate hidden motions with global motion aggregation.
\newblock In {\em ICCV}, pages 9772--9781, 2021.

\bibitem{realsense}
Leonid Keselman, John Iselin~Woodfill, Anders Grunnet-Jepsen, and Achintya
  Bhowmik.
\newblock Intel realsense stereoscopic depth cameras.
\newblock In {\em CVPR Workshops}, pages 1--10, 2017.

\bibitem{mpvit}
Youngwan Lee, Jonghee Kim, Jeffrey Willette, and Sung~Ju Hwang.
\newblock Mpvit: Multi-path vision transformer for dense prediction.
\newblock In {\em CVPR}, 2022.

\bibitem{CREStereo}
Jiankun Li, Peisen Wang, Pengfei Xiong, Tao Cai, Ziwei Yan, Lei Yang, Jiangyu
  Liu, Haoqiang Fan, and Shuaicheng Liu.
\newblock Practical stereo matching via cascaded recurrent network with
  adaptive correlation.
\newblock {\em ArXiv preprint arXiv:2203.11483}, 2022.

\bibitem{depthformer}
Zhenyu Li, Zehui Chen, Xianming Liu, and Junjun Jiang.
\newblock Depthformer: Exploiting long-range correlation and local information
  for accurate monocular depth estimation.
\newblock {\em ArXiv preprint arXiv:2203.14211}, 2022.

\bibitem{sttr}
Zhaoshuo Li, Xingtong Liu, Nathan Drenkow, Andy Ding, Francis~X Creighton,
  Russell~H Taylor, and Mathias Unberath.
\newblock Revisiting stereo depth estimation from a sequence-to-sequence
  perspective with transformers.
\newblock In {\em ICCV}, pages 6197--6206, 2021.

\bibitem{DySPN}
Yuankai Lin, Tao Cheng, Qi Zhong, Wending Zhou, and Hua Yang.
\newblock Dynamic spatial propagation network for depth completion.
\newblock In {\em AAAI}, 2022.

\bibitem{FCFRNet}
Lina Liu, Xibin Song, Xiaoyang Lyu, Junwei Diao, Mengmeng Wang, Yong Liu, and
  Liangjun Zhang.
\newblock Fcfr-net: Feature fusion based coarse- to-fine residual learning for
  depth completion.
\newblock In {\em AAAI}, volume~35, pages 2136--2144, 2021.

\bibitem{spn}
Sifei Liu, Shalini De~Mello, Jinwei Gu, Guangyu Zhong, Ming-Hsuan Yang, and Jan
  Kautz.
\newblock Learning affinity via spatial propagation networks.
\newblock {\em NeurIPS}, 2017.

\bibitem{swin}
Ze Liu, Yutong Lin, Yue Cao, Han Hu, Yixuan Wei, Zheng Zhang, Stephen Lin, and
  Baining Guo.
\newblock Swin transformer: Hierarchical vision transformer using shifted
  windows.
\newblock In {\em ICCV}, pages 10012--10022, 2021.

\bibitem{adamw}
Ilya Loshchilov and Frank Hutter.
\newblock Decoupled weight decay regularization.
\newblock {\em arXiv preprint arXiv:1711.05101}, 2017.

\bibitem{ma2019self}
Fangchang Ma, Guilherme~Venturelli Cavalheiro, and Sertac Karaman.
\newblock Self-supervised sparse-to-dense: Self-supervised depth completion
  from lidar and monocular camera.
\newblock In {\em ICRA}, pages 3288--3295. IEEE, 2019.

\bibitem{Ma2017SparseToDense}
Fangchang Ma and Sertac Karaman.
\newblock Sparse-to-dense: Depth prediction from sparse depth samples and a
  single image.
\newblock In {\em ICRA}, 2018.

\bibitem{kinect}
Microsoft.
\newblock Kinect for windows.
\newblock \url{https://developer.microsoft.com/en-us/windows/kinect/}.

\bibitem{semattnet}
Danish Nazir, Marcus Liwicki, Didier Stricker, and Muhammad~Zeshan Afzal.
\newblock Semattnet: Towards attention-based semantic aware guided depth
  completion.
\newblock {\em arXiv preprint arXiv:2204.13635}, 2022.

\bibitem{integrationvit}
Xuran Pan, Chunjiang Ge, Rui Lu, Shiji Song, Guanfu Chen, Zeyi Huang, and Gao
  Huang.
\newblock On the integration of self-attention and convolution.
\newblock In {\em CVPR}, pages 815--825, 2022.

\bibitem{NLSPN}
Jinsun Park, Kyungdon Joo, Zhe Hu, Chi-Kuei Liu, and In So~Kweon.
\newblock Non-local spatial propagation network for depth completion.
\newblock In {\em ECCV}, pages 120--136. Springer, 2020.

\bibitem{pytorch}
Adam Paszke, Sam Gross, Francisco Massa, Adam Lerer, James Bradbury, Gregory
  Chanan, Trevor Killeen, Zeming Lin, Natalia Gimelshein, Luca Antiga, Alban
  Desmaison, Andreas Kopf, Edward Yang, Zachary DeVito, Martin Raison, Alykhan
  Tejani, Sasank Chilamkurthy, Benoit Steiner, Lu Fang, Junjie Bai, and Soumith
  Chintala.
\newblock Pytorch: An imperative style, high-performance deep learning library.
\newblock In H. Wallach, H. Larochelle, A. Beygelzimer, F. d\textquotesingle
  Alch\'{e}-Buc, E. Fox, and R. Garnett, editors, {\em NeurIPS}, pages
  8024--8035. Curran Associates, Inc., 2019.

\bibitem{ConFormer}
Zhiliang Peng, Wei Huang, Shanzhi Gu, Lingxi Xie, Yaowei Wang, Jianbin Jiao,
  and Qixiang Ye.
\newblock Conformer: Local features coupling global representations for visual
  recognition.
\newblock In {\em ICCV}, pages 367--376, October 2021.

\bibitem{DeepLiDAR}
Jiaxiong Qiu, Zhaopeng Cui, Yinda Zhang, Xingdi Zhang, Shuaicheng Liu, Bing
  Zeng, and Marc Pollefeys.
\newblock Deeplidar: Deep surface normal guided depth prediction for outdoor
  scene from sparse lidar data and single color image.
\newblock In {\em CVPR}, pages 3313--3322, 2019.

\bibitem{dpt}
Ren\'{e} Ranftl, Alexey Bochkovskiy, and Vladlen Koltun.
\newblock Vision transformers for dense prediction.
\newblock {\em ArXiv preprint}, 2021.

\bibitem{GuideFormer}
Kyeongha Rho, Jinsung Ha, and Youngjung Kim.
\newblock Guideformer: Transformers for image guided depth completion.
\newblock In {\em CVPR}, pages 6250--6259, 2022.

\bibitem{mobilenetv2}
Mark Sandler, Andrew Howard, Menglong Zhu, Andrey Zhmoginov, and Liang-Chieh
  Chen.
\newblock Mobilenetv2: Inverted residuals and linear bottlenecks.
\newblock In {\em CVPR}, pages 4510--4520, 2018.

\bibitem{NYUV2}
Nathan Silberman, Derek Hoiem, Pushmeet Kohli, and Rob Fergus.
\newblock Indoor segmentation and support inference from rgbd images.
\newblock In {\em ECCV}, pages 746--760. Springer, 2012.

\bibitem{craft}
Xiuchao Sui, Shaohua Li, Xue Geng, Yan Wu, Xinxing Xu, Yong Liu, Rick Siow~Mong
  Goh, and Hongyuan Zhu.
\newblock Craft: Cross-attentional flow transformers for robust optical flow.
\newblock In {\em CVPR}, 2022.

\bibitem{GuideNet}
Jie Tang, Fei-Peng Tian, Wei Feng, Jian Li, and Ping Tan.
\newblock Learning guided convolutional network for depth completion.
\newblock {\em IEEE TIP}, pages 1116--1129, 2020.

\bibitem{KITTI}
Jonas Uhrig, Nick Schneider, Lukas Schneider, Uwe Franke, Thomas Brox, and
  Andreas Geiger.
\newblock Sparsity invariant cnns.
\newblock In {\em 3DV}, pages 11--20. IEEE, 2017.

\bibitem{FusionNet}
Wouter Van~Gansbeke, Davy Neven, Bert De~Brabandere, and Luc Van~Gool.
\newblock Sparse and noisy lidar completion with rgb guidance and uncertainty.
\newblock In {\em MVA}, pages 1--6. IEEE, 2019.

\bibitem{nlp}
Ashish Vaswani, Noam Shazeer, Niki Parmar, Jakob Uszkoreit, Llion Jones,
  Aidan~N Gomez, {\L}ukasz Kaiser, and Illia Polosukhin.
\newblock Attention is all you need.
\newblock {\em NeurIPS}, 2017.

\bibitem{pvt}
Wenhai Wang, Enze Xie, Xiang Li, Deng-Ping Fan, Kaitao Song, Ding Liang, Tong
  Lu, Ping Luo, and Ling Shao.
\newblock Pyramid vision transformer: A versatile backbone for dense prediction
  without convolutions.
\newblock In {\em ICCV}, pages 568--578, 2021.

\bibitem{cbam}
Sanghyun Woo, Jongchan Park, Joon-Young Lee, and In~So Kweon.
\newblock Cbam: Convolutional block attention module.
\newblock In {\em ECCV}, pages 3--19, 2018.

\bibitem{segformer}
Enze Xie, Wenhai Wang, Zhiding Yu, Anima Anandkumar, Jose~M Alvarez, and Ping
  Luo.
\newblock Segformer: Simple and efficient design for semantic segmentation with
  transformers.
\newblock {\em NeurIPS}, 34, 2021.

\bibitem{xiong2020sparse}
Xin Xiong, Haipeng Xiong, Ke Xian, Chen Zhao, Zhiguo Cao, and Xin Li.
\newblock Sparse-to-dense depth completion revisited: Sampling strategy and
  graph construction.
\newblock In {\em ECCV}, pages 682--699. Springer, 2020.

\bibitem{coat}
Weijian Xu, Yifan Xu, Tyler Chang, and Zhuowen Tu.
\newblock Co-scale conv-attentional image transformers.
\newblock In {\em ICCV}, pages 9981--9990, 2021.

\bibitem{DSPN}
Zheyuan Xu, Hongche Yin, and Jian Yao.
\newblock Deformable spatial propagation networks for depth completion.
\newblock In {\em ICIP}, pages 913--917. IEEE, 2020.

\bibitem{RigNet}
Zhiqiang Yan, Kun Wang, Xiang Li, Zhenyu Zhang, Baobei Xu, Jun Li, and Jian
  Yang.
\newblock Rignet: Repetitive image guided network for depth completion.
\newblock {\em ECCV}, 2022.

\bibitem{singlergbd}
Yinda Zhang and Thomas Funkhouser.
\newblock Deep depth completion of a single rgb-d image.
\newblock In {\em CVPR}, pages 175--185, 2018.

\bibitem{neighborattention}
Yongchi Zhang, Ping Wei, Huan Li, and Nanning Zheng.
\newblock Multiscale adaptation fusion networks for depth completion.
\newblock In {\em IJCNN}, pages 1--7, 2020.

\bibitem{monovit}
Chaoqiang Zhao, Youmin Zhang, Matteo Poggi, Fabio Tosi, Xianda Guo, Zheng Zhu,
  Guan Huang, Yang Tang, and Stefano Mattoccia.
\newblock Monovit: Self-supervised monocular depth estimation with a vision
  transformer.
\newblock In {\em 3DV}, 2022.

\bibitem{ACMNet}
Shanshan Zhao, Mingming Gong, Huan Fu, and Dacheng Tao.
\newblock Adaptive context-aware multi-modal network for depth completion.
\newblock {\em TIP}, pages 5264--5276, 2021.

\bibitem{zhong2019deep}
Yiqi Zhong, Cho-Ying Wu, Suya You, and Ulrich Neumann.
\newblock Deep rgb-d canonical correlation analysis for sparse depth
  completion.
\newblock {\em NeurIPS}, 2019.

\bibitem{RobustDC}
Yufan Zhu, Weisheng Dong, Leida Li, Jinjian Wu, Xin Li, and Guangming Shi.
\newblock Robust depth completion with uncertainty-driven loss functions.
\newblock In {\em AAAI}, 2022.

\end{thebibliography}
}

\end{document}